\documentclass[journal]{IEEEtran}
\IEEEoverridecommandlockouts

\usepackage{comment}
\usepackage{caption}
\usepackage{stfloats} 
\usepackage{booktabs}
\usepackage{diagbox}
\usepackage{tabularx}
\usepackage{multirow} 

\usepackage[linesnumbered,ruled]{algorithm2e} 

\SetAlFnt{\small}
\SetAlCapFnt{\small}
\SetAlCapNameFnt{\small}
\SetAlCapHSkip{0pt}
\IncMargin{-\parindent}
\newtheorem{definition}{Definition}

\usepackage{varwidth}
\usepackage{amsmath,amssymb,amsfonts,graphicx}
\usepackage{hyperref}
\hypersetup{colorlinks=true,citecolor = blue}
\usepackage{amssymb} 
\usepackage{array}    
\usepackage{arydshln}
\usepackage{graphics}
\usepackage{color}   
\usepackage{graphicx}  
\usepackage{graphicx,epstopdf,caption,url}  
\usepackage{algorithmic}
\usepackage[switch]{lineno}  
\begin{document}

\title{Graph Diffusion Network for Drug-Gene Prediction}
\author{
    Jiayang Wu, Wensheng Gan*, Philip S. Yu,~\IEEEmembership{Life Fellow,~IEEE} \\
    
    \thanks{This research was supported in part by National Natural Science Foundation of China (No. 62272196), Guangzhou Basic and Applied Basic Research Foundation (No. 2024A04J9971), and the Young Scholar Program of Pazhou Lab (No. PZL2021KF0023). (Corresponding author: Wensheng Gan)}
    
    \thanks{Jiayang Wu is with the College of Cyber Security, Jinan University, Guangzhou 510632, China. (E-mail: csjywu1@gmail.com)} 

    \thanks{Wensheng Gan is with the College of Cyber Security, Jinan University, Guangzhou 510632, China; and with Pazhou Lab, Guangzhou 510330, China. (E-mail: wsgan001@gmail.com)} 
    
    \thanks{Philip S. Yu is with the Department of Computer Science, University of Illinois Chicago, Chicago, IL 60607, USA. (E-mail: psyu@uic.edu)}
}

\maketitle

\begin{abstract}
    Predicting drug-gene associations is crucial for drug development and disease treatment. While graph neural networks (GNN) have shown effectiveness in this task, they face challenges with data sparsity and efficient contrastive learning implementation. We introduce a graph diffusion network for drug-gene prediction (GDNDGP), a framework that addresses these limitations through two key innovations. First, it employs meta-path-based homogeneous graph learning to capture drug-drug and gene-gene relationships, ensuring similar entities share embedding spaces. Second, it incorporates a parallel diffusion network that generates hard negative samples during training, eliminating the need for exhaustive negative sample retrieval. Our model achieves superior performance on the DGIdb 4.0 dataset and demonstrates strong generalization capability on tripartite drug-gene-disease networks. Results show significant improvements over existing methods in drug-gene prediction tasks, particularly in handling complex heterogeneous relationships. The source code is publicly available at \url{https://github.com/csjywu1/GDNDGP}.
\end{abstract}


\IEEEpeerreviewmaketitle

\section{Introduction}  \label{sec: introduction}

The application of artificial intelligence (AI) in bioinformatics has greatly advanced tasks like predicting protein-protein interactions (PPI) \cite{bryant2022improved}, gene regulatory network (GRN) \cite{pratapa2020benchmarking}, metabolic pathway \cite{baranwal2020deep}, and disease-disease associations (DDA) \cite{yu2021predicting}. These tasks are essential for understanding cellular processes \cite{ansari2012detecting}, metabolic functions \cite{dias2016genome}, and disease mechanisms \cite{luo2018ntshmda}, contributing to drug discovery and personalized medicine. AI-driven methods, particularly graph neural network (GNN) \cite{gu2022redda}, have proven effective in modeling complex biological networks, allowing for more accurate predictions of interactions between biological entities. However, the drug-gene prediction task remains challenging, which is crucial for drug development and therapeutic research. 

Drug-gene prediction plays a pivotal role in bioinformatics, aiming to identify interactions between drugs and genes. Such interactions are essential for advancing drug development \cite{harrison2023challenges}, pinpointing therapeutic targets \cite{pun2023ai}, and unraveling disease mechanisms at the genetic level \cite{cerro2023understanding}. Drug-gene interactions can reveal how specific drugs influence gene expression or interact with genetic variants, which is pivotal for designing personalized medicine approaches and repurposing existing drugs for new therapeutic uses. The importance of drug-gene prediction stems from its ability to improve the precision and efficiency of the drug-discovery process. By understanding how drugs affect gene networks, researchers can better predict drug efficacy, potential side effects, and interactions with genetic disorders. This information is invaluable for developing targeted therapies that are more effective and have fewer adverse effects. Moreover, drug-gene prediction plays a role in drug repositioning, where known drugs are found to have therapeutic effects on diseases outside their original intended use, thereby accelerating the drug development pipeline. AI-driven models, particularly those based on GNN, offer promising solutions to the challenges of drug-gene prediction \cite{gu2022redda}. Such models are capable of capturing the intricate relationships and dependencies between drugs and genes within heterogeneous biological networks. By learning meaningful representations of the interactions between these biological entities, GNN-based methods can improve the accuracy of predictions, leading to more reliable insights into drug mechanisms and potential therapeutic applications.

GNN-based methods are useful in drug-gene prediction. However, predicting relationships in drug-gene networks still poses additional challenges, such as data sparsity \cite{abbas2021application} and long-tail distributions \cite{baek2020learning}. For example, many drugs may have few known associations with genes, making it harder to generalize predictions based on limited information. To address these challenges, one promising solution is the construction of meta-paths for homogeneous message passing \cite{sun2011pathsim}. Meta-paths capture the higher-order relationships between entities, such as drug-gene-drug (D-G-D) and gene-drug-dene (G-D-G) \cite{meng2015discovering}, thereby facilitating the discovery of indirect connections. This approach not only mitigates the aforementioned challenges but also extends the algorithm's capabilities to handle tripartite networks, such as drug-gene-disease networks. Previous methods often focused on binary relationships, limiting their effectiveness when extended to multiple entity types \cite{fan2021heterogeneous}. Our experiments demonstrate that the introduced meta-path message passing can perform well in a tripartite network. Contrastive learning is frequently employed in link prediction tasks to differentiate positive from negative samples \cite{chuang2020debiased,sinha2021d2c}.  However, the volume of negative samples makes it computationally impractical to analyze them all. An effective approach to tackle this challenge involves generating high-quality, hard negative samples as a substitute for retrieving all unlinked data. In our framework, we adopt a graph diffusion network to generate these hard negative samples \cite{nguyen2024diffusion}. Incorporating the diffusion process allows us to create increasingly challenging negative samples, enhancing the model's robustness and generalization. In addition, it can generate various hard degree samples with different steps.

To tackle these challenges, we introduce a model named graph diffusion network for drug-gene prediction (GDNDGP). Our framework not only constructs the meta-path to facilitate effective information exchange between homogeneous and heterogeneous nodes but also generates high-quality negative samples using a diffusion mechanism to enhance the model's discriminative power during training. Our contributions to the field are manifold, encompassing two key areas:

\begin{itemize}
    \item We introduce the meta-path construction strategy that enhances message passing by capturing intricate relationships. This approach improves information exchange within both homogeneous (e.g., drug-drug, gene-gene) and heterogeneous (drug-gene) networks, ensuring more accurate interaction predictions.
    
    \item We integrate a graph diffusion network into our framework, allowing the model to generate high-quality hard negative samples. This process eliminates the need to retrieve large sets of unconnected pairs, improving training efficiency, and enhancing the model’s ability to generalize by progressively introducing more challenging negative samples.
\end{itemize}

The rest of this paper is organized as follows: Section \ref{sec: preliminaries} covers the preliminaries and foundational knowledge. Our proposed method, GDNDGP, is elaborated in Section \ref{sec: algorithm}. The experimental setup and results are presented in Section \ref{sec: experiments}, and we conclude the paper in Section \ref{sec: conclusion}.

\section{Related Work} \label{sec: relatedwork}
\subsection{Traditional methods}

Traditional methods often fall short in capturing the dynamic and diverse interactions. This shortcoming has led to a growing interest in leveraging artificial intelligence (AI) to enhance drug-gene research. Identifying disease-associated long non-coding RNAs (lncRNAs) has been the focus of various models, highlighting the role of non-coding genetic elements in diseases. GAERF \cite{wu2021gaerf}, introduced by Wu \textit{et al.}, employs graph autoencoders (GAE) combined with random forest techniques for detecting disease-related lncRNAs. In contrast, LPLNS \cite{li2018predicting} utilizes label propagation with linear neighborhood similarity to predict novel lncRNA-disease associations, thereby deepening our understanding of lncRNA roles in disease. lncRNA prediction methods add an extra layer of genetic interaction data, which, when integrated with drug-gene and miRNA predictions, offers a comprehensive view of genetic influences on diseases. Moreover, predicting disease-associated miRNAs sheds light on how these small RNA molecules regulate genes, serving as a vital complement to drug-gene interaction prediction. SMALF \cite{liu2021smalf}, created by Liu \textit{et al.}, integrates latent features from the miRNA-disease association matrix with original features, utilizing XGBoost \cite{chen2016xgboost} to predict unknown miRNA-disease associations. Likewise, CNNMDA \cite{xuan2019inferring}, introduced by Xuan \textit{et al.}, employs network representation learning in conjunction with convolutional neural networks (CNN).
 
\subsection{GNN-based methods}

Recent progress in AI, especially with graph neural networks (GNNs), has demonstrated significant potential in uncovering drug-gene associations. GNNs excel at capturing patterns from graph-structured data, making them highly effective in modeling drug-gene relationships. These models facilitate the identification of potential interactions, which is valuable for drug repurposing and discovering new therapeutic targets. For example, SGCLDGA \cite{fan2024sgcldga} integrates GNNs with contrastive learning, using graph convolutional network (GCN)  \cite{zhang2019graph} to derive vector representations of drugs and genes from a bipartite graph. It employs singular value decomposition (SVD) to refine the graph and create multiple views. Further, it optimizes these representations with a contrastive loss function to effectively differentiate positive and negative samples. In addition, multiple GNN-based methods have been developed for predicting miRNA-disease associations. HGCNMDA \cite{li2019novel} combines GNNs with node2vec and GCN to jointly learn features of miRNAs and diseases, offering a comprehensive perspective by integrating diverse biological networks. LAGCN \cite{yu2021predicting} employs graph convolution along with attention mechanisms to learn and integrate embeddings for drugs and diseases from heterogeneous networks, effectively concentrating on crucial features to enhance prediction accuracy. Lastly, NIMCGCN \cite{li2020neural} applies GCNs to uncover hidden features of miRNAs and diseases from similarity networks, thereby improving the discovery of novel miRNA-disease associations by leveraging similarities in known associations. These miRNA-disease prediction models contribute an additional layer of insight into the genetic underpinnings of diseases, which can aid in developing more targeted therapies when combined with drug-gene interaction predictions.

\subsection{Meta-path based methods}

Meta-path-based methods are extensively applied in drug-disease association prediction due to their effectiveness in revealing latent relationships through intermediate nodes within heterogeneous networks. A meta-path, defined as a sequence of relations linking different node types, captures essential semantic information, thereby enhancing predictive accuracy. MGP-DDA, in studying drug-disease associations, constructs a tripartite network involving drugs, gene ontology (go) functions, and diseases, employing three specific meta-paths to capture their interrelations \cite{kawichai2021meta}. These meta-paths generate features based on path instance counts, which are then used for drug-disease association prediction, showing robust performance by retaining richer semantic information from intermediate nodes. Similarly, HSGCLRDA integrates meta-path aggregation in a drug-disease-protein heterogeneous network \cite{wang2024hierarchical}. By constructing both global and local feature graphs through meta-paths, HSGCLRDA can capture detailed structural and contextual information about the interactions between drugs, diseases, and proteins. This meta-path-based aggregation is combined with contrastive learning to optimize feature representations, leading to improved prediction accuracy. Both methods highlight the strength of meta-path-based techniques in enhancing prediction tasks by retaining and utilizing rich semantic information from heterogeneous networks. These approaches not only enable the discovery of novel drug-disease associations but also contribute to the broader understanding of the underlying biological mechanisms.

\section{Preliminaries}  \label{sec: preliminaries}

Given a set of drugs $D $, a set of genes $G $, and a set of known associations $\mathcal{A} \subseteq D \times G $, the goal of drug-gene prediction is to identify unknown associations between drugs and genes. Specifically, for a given drug $d_i \in D $, our task is to estimate the likelihood of its association with a gene $g_j \in G $ using known drug-gene relationships. We represent the drug-gene prediction problem using a graph $\mathcal{G} = (\mathcal{V}, \mathcal{E}) $, where the vertex set $\mathcal{V} = D \cup G $ includes both drugs and genes, and the edge set $\mathcal{E} \subseteq D \times G $ denotes existing drug-gene associations. Our objective is to infer missing edges in $\mathcal{E} $, which correspond to potential novel drug-gene associations.

\begin{figure}[ht]
    \centering
    \includegraphics[clip,scale=0.45]{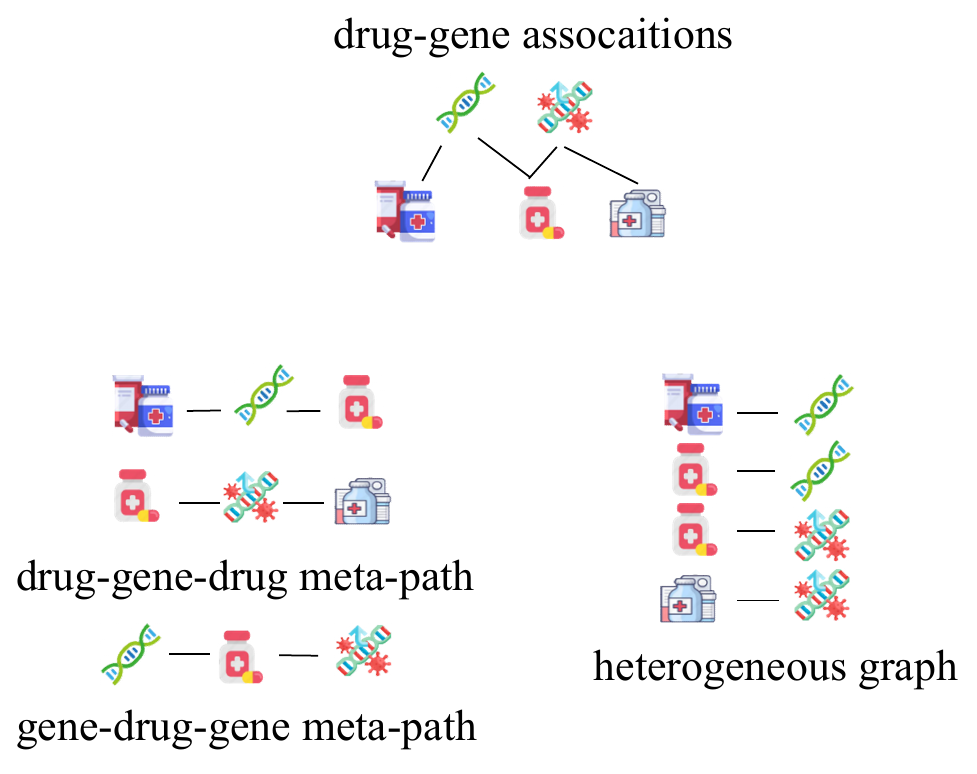}
    \caption{An illustration of drug-gene associations (top), meta-path construction (left), and heterogeneous graph message passing (right) used in our framework.}
    \label{fig:2}
\end{figure}

\begin{definition}[meta-path]   
    \rm  A meta-path is a sequence of node types and edge types that defines a composite relationship between two nodes in a heterogeneous graph \cite{zhang2014meta}. Specifically, in a graph where there are multiple types of nodes (e.g., drugs, genes, and diseases) and multiple types of relationships (e.g., drug-gene interactions, gene-disease associations), a meta-path describes a path schema that connects nodes through specific types of relationships. For instance, in a drug-gene prediction task, a meta-path such as drug-gene-drug (D-G-D) can be used to capture indirect associations between drugs by sharing common genes. Meta-paths are crucial for enhancing message passing in heterogeneous graphs, as they help capture higher-order relationships that would be missed by simple direct connections \cite{chang2022megnn,liang2022meta}. 
    In Fig. \ref{fig:2}, the meta-path is visually illustrated. The top section of the figure shows direct drug-gene associations, where drugs and genes are connected through known interactions. On the left, we see the construction of meta-paths that connect homogeneous nodes (such as drug-drug and gene-gene relationships) to capture more complex indirect relationships. On the right, the heterogeneous graph demonstrates how both drug and gene embeddings are updated through message passing. This integration of homogeneous and heterogeneous relationships helps improve the model’s ability to predict drug-gene interactions by leveraging both direct and indirect connections.
\end{definition}

\begin{definition}[homogeneous graph and heterogeneous graph]
    \rm A homogeneous graph consists of nodes and edges that all represent the same type of entity. The relationships between these entities are also of the same kind. For example, a graph where all nodes represent genes and all edges represent gene-gene interactions is considered a homogeneous graph. In contrast, a heterogeneous graph is a more complex structure that contains multiple types of nodes and edges. For instance, such a graph might include nodes for drugs and genes, with edges representing interactions like drug-gene associations.
\end{definition}

\begin{definition}[contrastive learning]
    \rm Contrastive learning is a self-supervised learning method designed to learn representations by distinguishing between positive and negative pairs \cite{wang2022contrastive}. In graph-based learning, contrastive learning encourages the model to assign higher similarity scores to positive pairs (e.g., drug-gene pairs with known associations) and lower similarity scores to negative pairs (e.g., drug-gene pairs without associations). Common loss functions in contrastive learning, such as InfoNCE \cite{parulekar2023infonce} or triplet loss \cite{yan2021beyond}, guide the model to develop meaningful embeddings by maximizing the similarity between positive samples while minimizing it for negative samples. This approach enhances the model's capability to differentiate between similar and dissimilar nodes within a graph, thereby improving its generalization performance.
\end{definition}

\begin{definition}[graph convolutional network (GCN)]
    \rm A GCN is a neural network architecture specifically designed to operate on graph-structured data \cite{zhang2019graph}. In GCNs, each node's feature representation is updated by aggregating information from its neighboring nodes. Formally, given a graph $\mathcal{G}$ = $(\mathcal{V}, \mathcal{E}) $ with node set $\mathcal{V} $ and edge set $\mathcal{E} $, and a feature matrix $\mathbf{X} \in \mathbb{R}^{N \times D} $ where $N $ denotes the number of nodes and $D $ represents the feature dimension, the forward propagation of a GCN layer is defined as:
    $\mathbf{H}^{(l+1)}$ = $\sigma \left( \tilde{\mathbf{D}}^{-\frac{1}{2}} \tilde{\mathbf{A}} \tilde{\mathbf{D}}^{-\frac{1}{2}} \mathbf{H}^{(l)} \mathbf{W}^{(l)} \right)$, where $\tilde{\mathbf{A}}$ = $\mathbf{A}$ + $\mathbf{I} $ is the adjacency matrix with self-loops added, $\tilde{\mathbf{D}} $ is the degree matrix of $\tilde{\mathbf{A}} $, $\mathbf{W}^{(l)} $ is the trainable weight matrix for the $l $-th layer, and $\sigma $ represents a non-linear activation function (e.g., ReLU). The initial input is $\mathbf{H}^{(0)}$ = $\mathbf{X} $, which is the original feature matrix. This layer-wise propagation allows GCNs to capture both the structural and feature information of the graph \cite{hamilton2017inductive}, making them effective for tasks like node classification, link prediction, and graph embedding.
\end{definition}

\begin{definition}[diffusion network]
    \rm A diffusion network is a model that simulates the process of diffusion across a graph, allowing information (or noise) to propagate over the nodes with time \cite{yang2023diffusion}. In the context of drug-gene prediction, a diffusion network can be employed to generate hard negative samples by gradually introducing noise into the gene embeddings and subsequently reversing this noise during training. During the forward process, noise is added to the gene embeddings across multiple time steps, resulting in a variety of negative samples. The reverse process then aims to recover the original gene embeddings by predicting and eliminating the noise. This diffusion mechanism facilitates the generation of hard negative samples for contrastive learning \cite{nichol2021improved}, enhancing the model's ability to differentiate between true and false associations in drug-gene interactions.
\end{definition}

\section{Algorithm} \label{sec: algorithm}

\begin{figure}[ht]
    \centering
    \includegraphics[clip,scale=0.5]{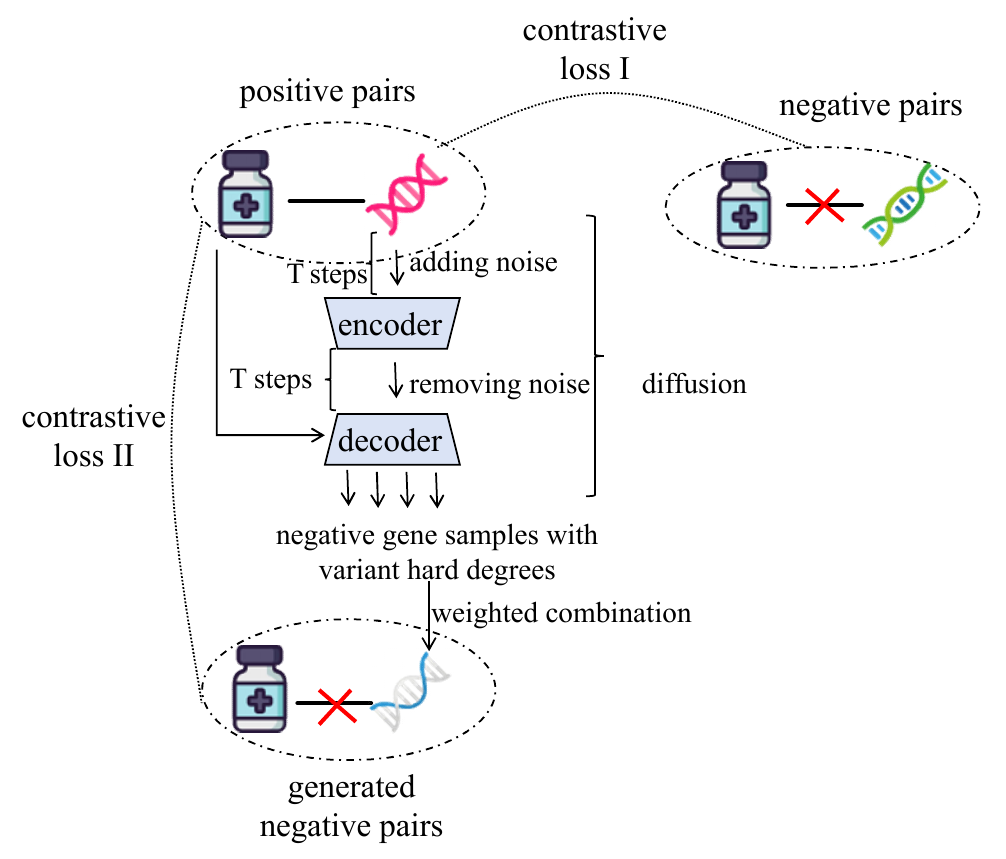}
    \caption{An illustration of the graph diffusion network in the GDNDGP. Positive drug-gene pairs are contrasted with two types of negative pairs: simple negative pairs generated from unlinked drug-gene pairs, and hard negative pairs generated through a diffusion process.}
    \label{fig:1}
\end{figure}

GDNDGP is a framework developed for drug-gene prediction utilizing GCN. Initially, it randomly initializes embeddings for both drugs and genes. To enhance the relationships among homogeneous nodes, the framework constructs meta-paths for drug-gene-drug and gene-drug-gene interactions, enabling efficient message passing not only between drugs and genes but also across drug-drug and gene-gene pairs. GDNDGP then facilitates message transfer within the heterogeneous graph, ensuring information exchange for both drugs and genes. To effectively distinguish the relationships between these entities, two types of contrastive learning losses are employed: contrastive loss I and contrastive loss II. Positive samples are derived from drug-gene pairs with confirmed associations, while negative samples are generated from unlinked drug-gene pairs. Due to the sheer number of negative samples, retrieving all hard negative examples presents a computational challenge. To address this, GDNDGP incorporates a graph diffusion network to generate hard negative samples, aiding in the contrastive learning process. As shown in Fig. \ref{fig:1}, the positive drug-gene pairs (top left) are contrasted with negative pairs (top right). The diffusion process (center) introduces noise to the negative samples through an encoder and removes noise through a decoder, generating progressively harder negative samples. These negative samples are then combined in a weighted manner, allowing the model to train more effectively. The final embeddings of drugs and genes are used to predict hidden associations between them, improving the accuracy of drug-gene predictions.

\subsection{Meta-path construction}

In meta-path construction, we focus on two types of meta-paths: drug-gene-drug (D-G-D) and gene-drug-dene (G-D-G). Let $D $ represent the set of drugs, and $G $ is the set of genes. For each drug $d \in D $, we first record all genes $G_d$ $\subseteq$ $G $ that are related to it. For drugs associated with multiple genes, we generate all possible gene-gene relationships under that drug. Specifically, for any drug $d \in D $ with an associated gene set $G_d$ = \{$g_1$, $g_2$, $\dots$, $g_n$\}, we form all possible gene pairs $(g_i, g_j) $, where $g_i$, $g_j$ $\in$ $G_d $ and $i \neq j $, thereby extracting all gene-gene relationships mediated by the drug. Similarly, for drug-drug relationships, we define two drugs $d_1$ $\in$ $D $, $d_2$ $\in$ $D $ as related if they share at least one common gene. This relation forms the drug-drug pair $(d_1, d_2) $ based on the shared gene information. To control the neighborhood size of each node (either drug or gene), we introduce a neighbor threshold, denoted as $\tau $. Let the set of neighbors for a node $v \in D \cup G $ be denoted as $\mathcal{N}(v) $. We limit the size of $\mathcal{N}(v) $ to at most $\tau $, i.e., $|\mathcal{N}(v)|$ $\leq $ $\tau $. If the number of neighbors of a node $v $ is less than $\tau $, we replicate its existing neighbors until $|\mathcal{N}(v)|$ = $\tau $. This ensures consistency in the neighborhood sizes across nodes and helps manage computational complexity, especially in the presence of large-scale relationships. The process is following from Algorithm \ref{alg:metapath}.

\begin{algorithm}
    \small
    \caption{meta-path construction}
    \label{alg:metapath}
    \begin{algorithmic}[1]
        \renewcommand{\algorithmicrequire}{\textbf{Input:}}
        \renewcommand{\algorithmicensure}{\textbf{Output:}}
        \REQUIRE{drug set $D$, gene set $G$, neighbor threshold $\tau$.}
        \ENSURE{gene-Gene relationships $\mathcal{R}_G$, drug-drug relationships $\mathcal{R}_D$}
        
        \STATE initialize $\mathcal{R}_G$ and $\mathcal{R}_D$ as empty sets;
        
        \FOR{each drug $d \in D$}
            \STATE $G_d \leftarrow$ all genes related to $d$;
            \FOR{each pair of genes $(g_i, g_j)$ $\in G_d$ with $i \neq j$}
                \STATE $\mathcal{R}_G \leftarrow \mathcal{R}_G $ $\cup$ $\{(g_i, g_j)\}$;
            \ENDFOR
        \ENDFOR

        \FOR{each gene $g \in G$}
            \STATE $D_g \leftarrow$ all drugs related to $g$;
            \FOR{each pair of drugs $(d_i, d_j)$ $\in D_g$ with $i \neq j$}
                \STATE $\mathcal{R}_D \leftarrow \mathcal{R}_D$ $\cup$ $\{(d_i, d_j)\}$;
            \ENDFOR
        \ENDFOR
        
        \FOR{each node $v \in D \cup G$}
            \STATE $N_v \leftarrow$ neighbors of node $v$ in $\mathcal{R}_D$ (if $v \in D$) or in $\mathcal{R}_G$ (if $v \in G$);
            \IF{$|N_v|$ $<$ $\tau$}
                \STATE replicate existing neighbors until $|N_v|$ = $\tau$;
            \ELSIF{$|N_v|$ $>$ $\tau$}
                \STATE randomly select $\tau$ neighbors from $N_v$;
            \ENDIF
        \ENDFOR
        
        \STATE \textbf{return:} gene-gene relationships $\mathcal{R}_G$ and drug-drug relationships $\mathcal{R}_D$.
    \end{algorithmic}
\end{algorithm}

\subsection{Graph aggregation network}

Details of the process of homogeneous graph generation. The relationships between drugs and genes are critical for defining the neighborhood structure, and these relationships are derived directly from the meta-paths (i.e., drug-gene-drug (D-G-D) and gene-drug-gene (G-D-G)). The input to the aggregation process consists of the drug set $ D $, the drug embeddings $ \mathbf{H}_D \in \mathbb{R}^{|D| \times d} $, the gene set $ G $, and the gene embeddings $ \mathbf{H}_G \in \mathbb{R}^{|G| \times d} $, where $ d $ is the embedding dimension. The meta-paths help define the relationships between drugs $ \mathcal{R}_D $ and genes $ \mathcal{R}_G $, which define the neighborhood structure for attention aggregation.

The D-G-D meta-path links drugs through shared genes, forming drug-drug relationships based on their common gene interactions. The G-D-G meta-path links genes through shared drugs, forming gene-gene relationships based on their common drug interactions. These meta-paths help to determine how neighbors should be selected and how much attention should be given to each neighbor during aggregation. By defining richer relationships in the graph, meta-paths enable the attention mechanism to focus on the most relevant neighbors—whether they are directly or indirectly connected through drugs and genes.

For each drug $ d_i$ $\in$ $D $, its neighbors are derived from the drug-gene relationships represented by $ \mathcal{R}_D $, which are constructed based on the meta-paths. For example, a drug-gene-drug meta-path will connect two drugs if they share a common gene. These relationships form the neighborhood $ \mathcal{N}(d_i) $ of drug $ d_i $. The attention mechanism is then applied to the neighbors of $ d_i $, which may include other drugs that are indirectly connected via common genes. The attention between $ d_i $ and its neighboring drugs $ d_j \in \mathcal{N}(d_i) $ is calculated by concatenating their embeddings, denoted as $ \mathbf{e}_{ij}$ = $[\mathbf{h}_{d_i} || \mathbf{h}_{d_j}] $, where $ || $ represents concatenation. The attention score $ e_{ij} $ is then computed as:
\begin{equation}
e_{ij} = \textit{LeakyReLU}(\mathbf{a}_{\Phi_m}^\top \mathbf{e}_{ij}),
\end{equation}
where $ \mathbf{a}_{\Phi_m} $ is the learnable attention vector. These attention scores are normalized using a softmax function to obtain the normalized attention weights:
\begin{equation}
\alpha_{ij} = \frac{\exp(e_{ij})}{\sum_{k \in \mathcal{N}(d_i)} \exp(e_{ik})}.
\end{equation}
Finally, the embedding of each drug $ d_i $ is updated through a weighted sum of its neighbors' embeddings:
\begin{equation}
\mathbf{h}'_{d_i} = \mathbf{h}_{d_i} + \sum_{d_j \in \mathcal{N}(d_i)} \alpha_{ij} \mathbf{h}_{d_j}.
\end{equation}

Similarly, for each gene $ g_i \in G $, the attention aggregation process follows the same procedure. The neighborhood for each gene $ g_i $ is determined by the relationships in $ \mathcal{R}_G $, which are also defined based on the meta-paths (e.g., gene-drug-gene (G-D-G)). The updated embedding for each gene $ g_i $ uses the same attention mechanism as for the drugs:
\begin{equation}
\mathbf{h}'_{g_i} = \mathbf{h}_{g_i} + \sum_{g_j \in \mathcal{N}(g_i)} \alpha_{ij} \mathbf{h}_{g_j}.
\end{equation}
This process captures the influence of connected drugs and genes based on their relationships in $\mathcal{R}_D $ and $\mathcal{R}_G $, as measured by the attention mechanism. By guiding the attention mechanism, the meta-paths ensure that the model focuses on the most relevant neighbors, facilitating the effective aggregation of information between connected drugs and genes.

Details of the process of heterogeneous graph generation. The drug embeddings $E_d^{(0)} $ and gene embeddings $E_g^{(0)} $, obtained from the homogeneous graph drug embeddings $H_d$ and gene embeddings $H_g$, are used as inputs. Specifically, the gene embeddings $E_g$ are updated based on the drug embeddings $E_d$, while drug embeddings $E_d$ are updated based on the gene embeddings $E_g$. The original adjacency matrix $\mathbf{A}$ represents the relationships between drugs and genes, where each entry $A_{ij} $ indicates the connection between drug $d_i$ and gene $g_j $. However, directly using the adjacency matrix $\mathbf{A}$ for graph-based operations can cause issues in terms of imbalance and numerical instability. Nodes with more connections (higher degree) would dominate the information propagation, leading to an uneven contribution from neighboring nodes. To address this, we use a normalized adjacency matrix $\mathbf{A}_{\textit{norm}} $, which ensures that the influence of each neighboring node is adjusted based on its degree \cite{chen2020iterative}. Specifically, the normalization ensures that contributions from neighbors are scaled by their degrees, preventing nodes with many connections from overwhelming nodes with fewer connections. This is achieved through symmetric normalization, where the degree matrix $\mathbf{D} $ is used to scale the adjacency matrix symmetrically, producing
$
\mathbf{A}_{\textit{norm}} = \mathbf{D}^{-1/2} \mathbf{A}' \mathbf{D}^{-1/2},
$
where $\mathbf{A}'$ = $\mathbf{A}$ + $\mathbf{I}$ is the adjacency matrix with added self-loops and $\mathbf{I}$ is the identity matrix. This symmetric normalization helps maintain balanced contributions from neighboring nodes during the aggregation process.

For gene embedding updates, the adjacency matrix $\mathbf{A} $ (representing the drug-gene relationship) is normalized and sparsely multiplied with the drug embeddings $E_d^{(0)} $ (the initial drug embedding). This operation is expressed as:
\begin{equation}
E_g = \mathbf{A}_{norm} E_d^{(0)},
\end{equation}
where $\mathbf{A}_{norm} $ represents the normalized adjacency matrix and $E_d^{(0)} $ is the drug embedding at the initial state. Similarly, to update the drug embeddings based on gene embeddings, the transposed normalized adjacency matrix $\mathbf{A}_{norm}^\top $ is multiplied with the initial gene embeddings $E_g^{(0)} $, represented as:
\begin{equation}
E_d = \mathbf{A}_{norm}^\top E_g^{(0)}.
\end{equation}
This process effectively propagates information between the drug and gene nodes, updating their embeddings in each step based on their mutual relationships. Sparse dropout is applied to the adjacency matrix to prevent overfitting. The process is following from Algorithm \ref{alg:graphagg}.

\begin{algorithm}
    \small
    \caption{Graph aggregation network}
    \label{alg:graphagg}
    \begin{algorithmic}[1]
        \renewcommand{\algorithmicrequire}{\textbf{Input:}}
        \renewcommand{\algorithmicensure}{\textbf{Output:}}
        \REQUIRE{A drug set $D$, a gene set $G$, relationships $\mathcal{R}_D$ and $\mathcal{R}_G$, an adjacency matrix $\mathbf{A}$, initial drug embeddings $H_d^{(0)}$, and initial gene embeddings $H_g^{(0)}$.}
        \ENSURE{updated drug embeddings $E_d$ and updated gene embeddings $E_g$.}
        
        \STATE \textbf{Step 1: attention-based aggregation for homogeneous drug and gene nodes}

        \FOR{each drug $d_i$ $\in D$ and neighbors $d_j$ $\in \mathcal{N}(d_i)$ from $\mathcal{R}_D$}
            \STATE compute attention score $e_{ij}$ and normalize using softmax to get $\alpha_{ij}$;
            \STATE update homogeneous drug embedding: $H_{d_i}'$ = $H_{d_i}$ + $\sum_{d_j} \alpha_{ij} H_{d_j}$;
        \ENDFOR
        
        \FOR{each gene $g_i$ $\in G$ and neighbors $g_j$ $\in \mathcal{N}(g_i)$ from $\mathcal{R}_G$}
            \STATE compute attention score $e_{ij}$ and normalize using softmax to get $\alpha_{ij}$;
            \STATE update homogeneous gene embedding: $H_{g_i}'$ = $H_{g_i}$ + $\sum_{g_j} \alpha_{ij} H_{g_j}$;
        \ENDFOR

        \STATE \textbf{Step 2: heterogeneous aggregation (drug-gene)}    
        \STATE initialize heterogeneous embeddings:
        $E_d^{(0)}$ = $H_d'$, $E_g^{(0)}$ = $H_g'$;

        \STATE compute normalized adjacency matrix: 
        $\mathbf{A}_{norm}$ = $\mathbf{D}^{-1/2} \mathbf{A} \mathbf{D}^{-1/2}$;
        
        \STATE update gene embeddings based on drug embeddings:
        $E_g$ = $\mathbf{A}_{norm} E_d^{(0)}$;
        
        \STATE update drug embeddings based on gene embeddings:
       $ E_d$ = $\mathbf{A}_{norm}^\top E_g^{(0)}$;

        \STATE \textbf{Return:} updated heterogeneous drug embeddings $E_d$, updated heterogeneous gene embeddings $E_g$.
    \end{algorithmic}
\end{algorithm}

\subsection{Graph diffusion network}

We use drug embedding as input to the graph diffusion network to generate negative gene embeddings. The graph diffusion network can be simply divided into a forward process and a backward process. In the forward process of the diffusion network, gene embedding is input, and negative samples are generated over $T$ time steps. For each time step $t \in \{1, 2, \dots, T\}$, the negative gene embedding $\mathbf{E}_{g_t}$ is calculated as follows:

\begin{equation}
    \mathbf{E}_{g_t} = \sqrt{\bar{\alpha}_t} \mathbf{E}_{g_0} + \sqrt{1 - \bar{\alpha}_t} \mathbf{\epsilon}_t.
\end{equation}

Here, $\mathbf{E}_{g_t}$ represents the negative gene embedding at time step $t$, and $\mathbf{E}_{g_0}$ is the initial gene embedding. The noise vector $\mathbf{\epsilon}_t$ is drawn from a standard normal distribution, $\mathbf{\epsilon}_t \sim \mathcal{N}(0, I)$, where $I$ is the identity matrix.

The coefficient $\alpha_t$ controls the amount of noise added at each time step, and it is defined as follows:

\begin{small}
    \begin{equation}
    \alpha_t = \alpha_{\textit{start}} + \left( \frac{t}{T-1} \right) \cdot \left( \alpha_{\textit{end}} - \alpha_{\textit{start}} \right), \quad t = 0, 1, 2, \dots, T-1.
\end{equation}
\end{small}

The cumulative product $\bar{\alpha}_t$ is the product of all $\alpha$ values up to the time step $t$, i.e.,  $\bar{\alpha}_t = \prod_{i=1}^{t} \alpha_i.$ The noise vectors $\mathbf{\epsilon}_t$ for each time step are independently sampled from a standard normal distribution. Throughout the diffusion process, noise $\mathbf{\epsilon}_t$ is progressively added at each time step, generating increasingly diverse negative gene embeddings. After $T$ time steps, the final negative gene embedding $\mathbf{E}_{g_T}$ is obtained, integrating the cumulative effect of noise $\{\mathbf{\epsilon}_1, \mathbf{\epsilon}_2, \dots, \mathbf{\epsilon}_T\}$ at each step.

In the backward process, the model aims to recover the original gene embedding by predicting the noise added during the forward process. The reverse process begins with a random noise vector $\mathbf{\epsilon}_T$ at time step $T$. Starting from this noisy state, the model progressively refines the noisy embedding by predicting the noise at each time step and removing it, gradually recovering the clean gene embedding. At each time step $t \in \{T, T-1, \dots, 1\}$, the model predicts the noise $\hat{\mathbf{\epsilon}}_\theta$ using the predicted gene embedding at the time step $t$, denoted as $\hat{\mathbf{E}}_{g_t}$, the drug embedding $\mathbf{E}_{d_0}$, and the time embedding $\textit{PE}(t)$. The noise prediction function is expressed as:

\begin{equation}
    \hat{\mathbf{\epsilon}}_\theta(\hat{\mathbf{E}}_{g_t}, \mathbf{E}_{d_0}, \textit{PE}(t)).
\end{equation}

The time embedding $\textit{PE}(t)$ is combined with the predicted gene embedding $\hat{\mathbf{E}}_{g_t}$ and the drug embedding $\mathbf{E}_{d_0}$ to predict the noise. The predicted noise $\hat{\mathbf{\epsilon}}_\theta(\cdot)$ is then used to refine the noisy embedding $\hat{\mathbf{E}}_{g_t}$ by gradually removing the noise at each time step.

The model calculates the mean $\mu_\theta(\hat{\mathbf{E}}_{g_t}, t)$ of the posterior distribution $q(\hat{\mathbf{E}}_{g_{t-1}} | \hat{\mathbf{E}}_{g_t}, \mathbf{E}_{d_0})$ as follows:

\begin{small}
\begin{equation}
    \widetilde{\mathbf{E}}_{g_{t-1}} = \mu_\theta(\hat{\mathbf{E}}_{g_t}, t) = \frac{1}{\sqrt{\alpha_t}} \left( \hat{\mathbf{E}}_{g_t} - \frac{1 - \bar{\alpha}_t}{\sqrt{1 - \bar{\alpha}_t}} \hat{\mathbf{\epsilon}}_\theta(\hat{\mathbf{E}}_{g_t}, \mathbf{E}_{d_0}, \textit{PE}(t)) \right).
\end{equation}
\end{small}

Then, $\hat{\mathbf{E}}_{g_{t-1}}$ is calculated by adding a noise component scaled by the posterior variance $\sigma_t^2$:
\begin{equation}
    \hat{\mathbf{E}}_{g_{t-1}} = \mu_\theta(\hat{\mathbf{E}}_{g_t}, t) + \sigma_t \mathbf{z} = \widetilde{\mathbf{E}}_{g_{t-1}} + \sigma_t \mathbf{z},
\end{equation}
where $\mathbf{z} \sim \mathcal{N}(0, \mathbf{I})$ is Gaussian noise. The posterior variance $\sigma_t^2$ is given by:

\begin{equation}
    \sigma_t^2 = \frac{1 - \bar{\alpha}_{t-1}}{1 - \bar{\alpha}_t}.
\end{equation}

The time embedding $\textit{PE}(t)$, which encodes temporal information, is computed as follows:
\begin{equation}
    \textit{PE}(t) = \left[ \sin\left(t \cdot \omega_i\right), \cos\left(t \cdot \omega_i\right) \right]_{i=1}^{\frac{D}{2}}.
\end{equation}
where the frequency term $\omega_i$ is defined as:

\begin{equation}
    \omega_i = \exp\left( -\frac{\log(10000)}{\frac{D}{2}-1} \cdot (i-1) \right).
\end{equation}

The time embedding $\textit{PE}(t)$ is combined with the predicted gene embedding $\hat{\mathbf{E}}_{g_t}$ and the drug embedding $\mathbf{E}_{d_0}$ to predict the noise. The loss function used to train the diffusion network measures the difference between the predicted noise $\hat{\mathbf{\epsilon}}_\theta(\hat{\mathbf{E}}_{g_t}, \mathbf{E}_{d_0}, \textit{PE}(t))$ and the true noise $\mathbf{\epsilon}_t$ from the forward process. This is calculated using the mean squared error loss:

\begin{equation}
    L = \mathbb{E}_{t, \hat{\mathbf{E}}_{g_t}, \mathbf{E}_{d_0}, \mathbf{\epsilon}_t} \left[ |\mathbf{\epsilon}_t - \hat{\mathbf{\epsilon}}_\theta(\hat{\mathbf{E}}_{g_t}, \mathbf{E}_{d_0}, \textit{PE}(t))|_2^2 \right].
\end{equation}

By minimizing this loss, the model learns to accurately predict the noise injected during the forward process. This enables the reverse diffusion process to progressively remove the noise from the noisy gene embeddings, ultimately recovering the original clean embedding.

During the backward diffusion process, the negative embeddings generated at each step are stored in a list $L_{\textit{neg}}$, which includes the progressively denoised embeddings from time step $T$ down to 1. This list is represented as:
\begin{equation}
    L_{\textit{neg}} = \{ \hat{\mathbf{E}}_{g_T}, \hat{\mathbf{E}}_{g_{T-1}}, \dots, \hat{\mathbf{E}}_{g_1} \}.
\end{equation}

Instead of using all embeddings, the model samples embeddings from specific fractional time steps $T/1$, $T/2$, $T/3$, and $T/4$, capturing embeddings at different levels of refinement. These sampled embeddings are denoted as:
\begin{equation}
    L_{\textit{sampled}} = \{ \hat{\mathbf{E}}_{g_{T/1}}, \hat{\mathbf{E}}_{g_{T/2}}, \hat{\mathbf{E}}_{g_{T/3}}, \hat{\mathbf{E}}_{g_{T/4}} \}.
\end{equation}

At this stage, the model applies a weight matrix $W$, where larger weights are assigned to embeddings from earlier time steps (i.e., higher noise levels). The rationale is that noisier embeddings retain more diverse information, which can improve the model’s robustness. The weight matrix $W$ is applied in descending order, such that:
\begin{equation}
    W_T > W_{T/2} > W_{T/3} > W_{T/4}.
\end{equation}

This ensures that embeddings with more noise, which carry more diverse information, are given greater importance. The final gene embedding $\hat{E}_g^{\textit{neg}}$ is then computed as a weighted sum of these sampled embeddings:
\begin{equation}
    \hat{E}_g^{\textit{neg}} = \sum_{i=1}^{4} W_{T/i} \cdot \hat{\mathbf{E}}_{g_{T/i}}.
\end{equation}

In this expression, $W_{T/i}$ represents the weight assigned to the embedding $\hat{\mathbf{E}}_{g_{T/i}} $, sampled at fractional time steps $T/1$, $T/2$, $T/3$, and $T/4$, with larger weights assigned to earlier, noisier embeddings. This approach ensures that the model captures valuable information from different stages of the backward diffusion process. The process of the graph diffusion network is shown in Algorithm \ref{alg:graphdiffusion}.

\begin{algorithm}
    \small
    \caption{graph diffusion network}
    \label{alg:graphdiffusion}
    \begin{algorithmic}[1]
        \renewcommand{\algorithmicrequire}{\textbf{Input:}}
        \renewcommand{\algorithmicensure}{\textbf{Output:}}
        \REQUIRE{drug embeddings $E_d^{(0)}$, gene embeddings $E_g^{(0)}$, noise schedule $\{\alpha_t\}$, total time steps $T$, and weight matrix $W$.}
        \ENSURE{final negative gene embedding $E_g^{neg}$.}
        
        \FOR{$t$ = 1 to $T$}
            \STATE sample noise $\epsilon_t \sim \mathcal{N}(0, I)$
            \STATE update gene embedding: $E_{g_t}$ = $\sqrt{\bar{\alpha}_t} E_{g_0} + \sqrt{1 - \bar{\alpha}_t} \epsilon_t$
        \ENDFOR
            
        \STATE initialize $\hat{E}_{g_T}$ as random noise
        \FOR{$t = T$ to $1$}
            \STATE predict noise: $\hat{\epsilon}_\theta(\hat{E}_{g_t}, E_{d_0}, \textit{PE}(t))$
            \STATE update gene embedding: $\hat{E}_{g_{t-1}}$ = $\mu_\theta(\hat{E}_{g_t}, t) + \sigma_t z$
         \ENDFOR
         \STATE store all negative embeddings $\hat{E}_{g_T}$, $\hat{E}_{g_{T-1}}$, $\dots$, $\hat{E}_{g_1}$ in $L_{\textit{neg}}$
        \STATE sample embeddings from $L_{\textit{neg}}$ at $T/1$, $T/2$, $T/3$, $T/4$
        \STATE apply weights: $E_g^{\textit{neg}}$ = $W_T \cdot E_{g_{T/1}} + W_{T/2}$ $\cdot E_{g_{T/2}}$ + $W_{T/3}$ $\cdot $ $E_{g_{T/3}}$ + $W_{T/4} \cdot $ $E_{g_{T/4}}$;
        
        \STATE \textbf{return:} $E_g^{neg} $
    \end{algorithmic}
\end{algorithm}

\subsection{Objective and optimization}

Our training objective consists of three components: the diffusion model loss $L_{\textit{diffusion}}$ and two contrastive losses. The diffusion model loss $L_{\textit{diffusion}}$ is used to train the model to predict the noise added during the forward diffusion process, enabling it to reconstruct the original gene embedding from its noisy version generated during diffusion. This process ensures that the model can effectively reverse the diffusion process and recover clean gene embeddings, as detailed in the previous section.

The first contrastive loss uses cross-entropy to encourage the model to assign higher scores to positive drug-gene pairs and lower scores to negative pairs. We use a multilayer perceptron \textit{MLP$_1$} that takes the concatenation of the drug embedding and the gene embedding as input. \textit{MLP$_1$} is a two-layer multilayer perceptron that takes the concatenation of the drug embedding and the gene embedding as input. The first layer applies a linear transformation followed by a ReLU activation function, and the second layer outputs the prediction. For positive pairs (i.e., linked drug and gene embedding), \textit{MLP$_1$} is trained to output a score close to 1, and for negative pairs (i.e., drug embeddings and real negative gene embedding), it outputs a score close to 0. The cross-entropy loss is expressed as:
\begin{small}
    \begin{equation}
L_{\textit{CE}} = -\log\sigma(\textit{MLP}_1(E_d, E_g)) - \log(1 - \sigma(\textit{MLP}_1(E_d, E_g^{\textit{neg}}))),
\end{equation}
\end{small}
where $\sigma$ is the sigmoid activation function, $E_d$ is the drug embedding, $E_g$ is the positive gene embedding, and $E_g^{\textit{neg}}$ is the real negative gene embedding.

The second contrastive loss is a margin-based ranking loss applied to the embeddings generated by the diffusion model. We introduce another multilayer perceptron network $\textit{MLP}_2$ to compute scores for drug-gene pairs with the same structure as $\textit{MLP}_1$. This loss encourages the score for positive pairs to be higher than that for negative pairs. The margin-based ranking loss is formulated as:
\begin{small}
    \begin{equation}
    L_{\textit{margin}} = \mathbb{E}_{E_d, E_g, \hat{E}_g^{\textit{neg}}} \left[ \max(0, \textit{MLP}_2(E_d, E_g) - \textit{MLP}_2(E_d, \hat{E}_g^{\textit{neg}})) \right],
\end{equation}
\end{small}
where$\hat{E}_g^{\textit{neg}}$ is the negative gene embedding generated by the diffusion model.

The total loss function combines these three components into the following:
\begin{small}
    \begin{equation}
L_{\textit{total}} = L_{\textit{diffusion}} + L_{\textit{CE}} + L_{\textit{margin}}.
\end{equation}
\end{small}
By minimizing $L_{\textit{total}}$, the model learns robust representations for both positive and negative drug-gene interactions.

During inference, the model comprises two contrasts: one is from the positive pairs and negatives from the graph, and the other is the positive pairs and negative pairs generated from the diffusion network. The diffusion network generates negative gene embeddings conditioned on the drug embedding. Negative embeddings are sampled from various timesteps in the reverse diffusion process (e.g., $T$, $T$-1, $T$-2, $T$-3) and are assigned different weights to capture diverse negative examples. These weighted embeddings are combined to form the final negative gene embedding $\hat{E}_g^{\textit{neg}}$

The MLP network computes scores for drug-gene pairs to predict their likelihood of interaction. The final score is calculated by combining the outputs from both MLPs as follows:
\begin{small}
    \begin{equation}
    \textit{Score} = \textit{MLP}_2(E_d^0, E_g^0) - \textit{MLP}_2(E_d^0, \hat{E}_g^{\textit{neg}}) + \textit{MLP}_1(E_d^0, E_g^0).
\end{equation}
\end{small}

This scoring mechanism ensures that positive drug-gene pairs receive higher scores compared to negative pairs, aligning with the training objective. By integrating the diffusion network with contrastive losses, the model effectively generates diverse negative samples and distinguishes between linked and unlinked drug-gene pairs, improving prediction performance for drug-gene interactions.

\subsection{Complexity analysis}
In this section, we analyze the time complexity of our framework, which uses a homogeneous graph approach, compared to a fully transformer-based model. The key distinction between these two approaches lies in how they handle the relationships between nodes during message passing.

Our framework utilizes a homogeneous graph where each node communicates only with its immediate neighbors. Additionally, it restricts the number of neighbors for each node to a fixed threshold $\tau $, ensuring that message aggregation occurs locally within a small, controlled set of nodes. The time complexity of message passing in the attention calculation is primarily influenced by the number of edges in the graph. For a graph with $N $ nodes and $E $ edges, each node aggregates information from its neighbors, resulting in a time complexity of $O(E) $. Given that we limit the number of neighbors to $\tau $, the complexity per node becomes $O(\tau) $, and for the entire graph, the total complexity is $O(N \cdot \tau) $. Since $\tau $ is typically a small constant, the overall time complexity remains linear w.r.t. the number of nodes, i.e., $O(N) $.

A transformer-based model considers interactions between all nodes, regardless of whether they are neighbors. In this approach, the attention mechanism computes relationships between every pair of nodes in the graph. As a result, the time complexity for a graph with $N $ nodes is $O(N^2) $, since each node interacts with every other node. This quadratic complexity makes transformers computationally expensive when applied to large graphs, as the number of interactions grows rapidly with the size of the graph.

When comparing the two approaches, the homogeneous graph-based GCN model is significantly more efficient. By focusing only on local interactions between a node and its immediate neighbors, and limiting the number of neighbors with the threshold $\tau $, the computation is constrained to $O(N) $, which is scalable even for large graphs. In contrast, the transformer-based model, with its $O(N^2) $ complexity, becomes much less efficient as the size of the graph increases, due to the need to process interactions between all nodes.

In summary, our framework, which uses a homogeneous graph structure, is computationally more efficient compared to transformer-based models, especially for large-scale graphs. By limiting interactions to local neighborhoods, we can significantly reduce the computational cost while maintaining effective message passing.

\section{Experiments} \label{sec: experiments}

Our experiments were carried out on the system equipped with the NVIDIA GeForce RTX 3080 GPU, using 24 GB memory. Detailed information about the experimental setup is provided below. The source code is publicly available at \url{https://github.com/csjywu1/GDNDGP}

\begin{itemize}
    \item \textbf{Q1 (Effectiveness):} How effective is GDNDGP compared to other baseline methods in predicting drug-gene associations?
    
    \item \textbf{Q2 (Ablation study):} What is the impact of removing or modifying key components of GDNDGP on its overall performance?
    
    \item \textbf{Q3 (Hyper-parameter analysis):} How do different hyperparameters affect the performance of GDNDGP?
    
    \item \textbf{Q4 (Time analysis):} How does the computational efficiency of GDNDGP compare in terms of training and inference time?
\end{itemize}

\subsection{Datasets}

The data used in this study comes from the Drug–Gene Interaction Database 4.0 (DGIdb 4.0) \cite{freshour2021integration}, which is a comprehensive resource providing detailed information on known drug-gene associations. DGIdb 4.0 includes 54,591 drug-gene interactions, involving 41,102 genes and 14,449 drugs. For our experiments, we followed the data preparation method used in SGCLDGA \cite{fan2024sgcldga}, which leverages 46,892 established drug-gene associations from DGIdb 4.0, including 10,690 drugs and 3,227 genes. To create negative samples, we generated two unlinked drug-gene pairs for each drug, resulting in a total of 21,380 negative samples. The dataset was then split into training and test sets with an 8:2 ratio. To further validate the generalizability of our framework, we conducted experiments on a drug-go-disease tripartite network \cite{kawichai2021meta}, which contains drug-go associations and go-disease associations. We need to predict the drug-diseases through the meta-paths drug-go and disease-go. Table \ref{table:summary} is the statistics for the two datasets.

\begin{table}[ht]
    \centering
    \caption{Statistics of the DGIdb 4.0 and drug-go-disease tripartite datasets}
    \label{table:summary}
    \renewcommand{\arraystretch}{1.5} 
    \begin{tabular}{c|c|c|c}
    \hline
    \textbf{Dataset} & \textbf{Type}                  & \textbf{Category} & \textbf{Number} \\ \hline
    \multirow{3}{*}{DGIdb 4.0} & nodes                   & drugs             & 10,690          \\ \cline{3-4} 
                                &                        & genes             & 3,227           \\ \cline{2-4} 
                                & known interactions     & drug-gene         & 46,892          \\ \hline
    \multirow{6}{*}{drug-go-disease} & \multirow{3}{*}{nodes}  & drugs             & 1,022           \\ \cline{3-4} 
                                     &                        & diseases          & 585             \\ \cline{3-4} 
                                     &                        & go terms          & 8,320           \\ \cline{2-4} 
                                     & \multirow{3}{*}{known interactions} & drug-disease      & 6,710           \\ \cline{3-4} 
                                     &                        & drug-go           & 52,463          \\ \cline{3-4} 
                                     &                        & disease-go        & 92,135          \\ \hline
\end{tabular}
\end{table}

\subsection{Baselines and experimental settings}

To assess the performance of GDNDGP, we employed a 5-fold cross-validation (5-fold CV) approach on our experimental dataset. In each iteration of the cross-validation, four subsets were used to train the model, while the remaining subset was used for testing. We utilized five commonly used metrics to evaluate the model performance: AUC, AUPR, Recall, Precision, and F1-score. The formulas for computing Recall, Precision, and F1-score \cite{wu2021gaerf}.
The following metrics  MCC, Spec. and NPV \cite{zhou2019deep} were also used to evaluate the model: 
\begin{equation} \label{eq33}
    \textit{MCC} = \frac{TP \times TN - FP \times FN}{\sqrt{(TP + FP)(TP + FN)(TN + FP)(TN + FN)}},
\end{equation}

\begin{equation} \label{eq34}
    \textit{Spec.} = \frac{TN}{TN + FP},
\end{equation}

\begin{equation} \label{eq35}
    \textit{NPV} = \frac{TN}{TN + FN},
\end{equation}

To validate the effectiveness of GDNDGP, we carefully selected eight state-of-the-art models for comparison, representing both GNN-based and meta-path-based approaches. The GNN-based methods were chosen to evaluate different aspects of graph learning capabilities, ranging from simplified architectures (LightGCN) to attention mechanisms (LAGCN) and contrastive learning approaches (SGCLDGA). These methods provide benchmarks for assessing GDNDGP's graph representation learning and negative sampling strategies. The meta-path-based methods were selected to specifically evaluate GDNDGP's effectiveness in handling complex heterogeneous networks and meta-path learning, particularly important for drug-gene-disease predictions. This diverse set of baselines enables a comprehensive evaluation of our model's performance across different technical approaches.

\textbf{GNN-based methods}: \textbf{LightGCN} \cite{he2020lightgcn}: A simplified GCN model that eliminates unnecessary complexity while focusing on essential graph convolution operations.  \textbf{LAGCN} \cite{yu2021predicting}: Incorporates an attention mechanism to enhance link prediction performance by selectively focusing on important graph connections. \textbf{MF} \cite{lee2000algorithms}: A classic matrix factorization method used for predicting drug-gene interactions by approximating the interaction matrix.  \textbf{AGAEMD} \cite{zhang2022predicting}: An autoencoder-based approach designed for drug-gene prediction, utilizing representation learning to capture underlying features.  \textbf{MNGACDA} \cite{yang2023predicting}: A GCN model that utilizes multi-modal data for more comprehensive drug-gene association prediction. \textbf{SGCLDGA} \cite{fan2024sgcldga}: It uses self-supervised contrastive learning to distinguish between positive and negative drug-gene associations.

\textbf{Meta-path-based methods}: \textbf{HSGCLRDA} \cite{wang2024hierarchical}: A hierarchical graph-based model that leverages meta-path information and multi-hop relationships in a heterogeneous network to predict drug-gene interactions. \textbf{MGP-DDA} \cite{kawichai2021meta}: A method based on meta-path gene ontology profiles for predicting drug-disease associations, which helps in understanding drug-go-disease relationships.

By comparing GDNDGP with these models, we aim to demonstrate its competitiveness in both graph neural network-based approaches and meta-based methods. The experimental settings were standardized across all models to ensure a fair comparison, and the same 5-fold CV method was applied to each model. The hyperparameters used in GDNDGP are as follows: The batch size is 400, balancing memory efficiency and sample coverage during training. The embedding dimension is fixed at 128, which strikes a balance between accuracy and computational efficiency. The learning rate (lr) is set to 0.001 to ensure stable convergence with the Adam optimizer. The neighbor threshold is 30, limiting each node's connections for efficient computation. Additionally, both homogeneous and heterogeneous graphs use a 1-layer network architecture, ensuring a consistent structural configuration across different types of graphs. The number of diffusion steps $T$ is set to 100 to introduce sufficient noise in the forward process. For the noise schedule, $\alpha_t$ values are predefined to gradually reduce noise during the diffusion process. At the initial step ($t$ = 0), $\alpha_0$ is set to 0.9999, representing minimal noise, and it progressively decreases to $\alpha_T$ = 0.98 at the final time step ($t$ = $T$), ensuring a smooth transition of noise over the diffusion steps. During the reverse process, different weights, represented by the weight matrix $W$ (i.e., $W_T$ = 0.9, $W_{T/2}$ = 0.8, and $W_{T/3} $ = 0.7, $W_{T/4}$ = 0.6), are assigned to the embeddings sampled from various diffusion steps. This weighted sampling ensures that diverse negative examples are captured, enhancing the model’s generalization capabilities.

\subsection{Effectiveness (Q1)}

To evaluate the effectiveness of the GDNDGP model, we compared its performance against other GNN-based and meta-path-based methods across two datasets: DGIdb 4.0 and drug-go-disease. The evaluation metrics include AUC, AUPR, Recall, Precision, F1-score, MCC, Specificity, and NPV, as shown in Table \ref{table:performance1} and Table \ref{table:performance2}. For the DGIdb 4.0 dataset, GDNDGP achieves an AUC of 0.9531, representing a significant improvement of 17.3\% over LightGCN (0.8128) and showing a 6.8\% increase over SGCLDGA (0.8921). Additionally, GDNDGP demonstrates a higher AUPR (0.9816) and F1-score (0.9310), reflecting superior performance in differentiating positive and negative samples and improving overall classification accuracy. Notably, GDNDGP achieves an MCC of 0.9045, substantially higher than both GNN-based methods like SGCLDGA (0.8692) and other meta-path-based approaches like HSGCLRDA (0.8635), indicating a stronger correlation between predicted and actual classifications. The model also demonstrates robust performance in identifying true negatives, with a Specificity of 0.9108 and NPV of 0.8966, outperforming all baseline models in these metrics.
When evaluating the drug-go-disease dataset, GDNDGP attains an AUC of 0.9689, marking a 30.9\% improvement over LAGCN (0.7403) and an 18.1\% increase over SGCLDGA (0.8201). The significant performance gains highlight GDNDGP's ability to effectively handle complex datasets like drug-go-disease. Improvements are also observed across Recall (0.9166), Precision (0.9157), and F1-score (0.9386), further indicating GDNDGP's capacity to capture intricate relationships in multi-hop associations more accurately than other models. The model's superior performance is further validated by its MCC score of 0.9127, significantly higher than SGCLDGA (0.8543) and HSGCLRDA (0.8932). GDNDGP also achieves the highest Specificity (0.9256) and NPV (0.9189) among all compared methods, demonstrating its exceptional ability to correctly identify true negative cases and maintain high prediction reliability for negative instances in complex heterogeneous networks.
Fig. \ref{fig:graphs1} illustrates the ROC-AUC comparison between GDNDGP with the two best-performing baselines SGCLDGA (from GNN-based), and HSGCLRDA (from meta-path based) on two datasets. For the DGIdb 4.0 dataset, GDNDGP (green curve) outperforms SGCLDGA (AUC = 0.8921) and HSGCLRDA (AUC = 0.9340), achieving an AUC of 0.9531. This result underscores GDNDGP's superior predictive power in drug-gene association tasks, particularly in simpler datasets, where GNN-based methods like SGCLDGA are less accurate.
In the drug-go-disease dataset, GDNDGP's ROC curve (green) demonstrates a larger area under the curve, achieving an AUC of 0.9689. SGCLDGA (blue, AUC = 0.8201) performs worse, while HSGCLRDA (red, AUC = 0.9243) approaches but doesn't quite match GDNDGP's performance. This showcases GDNDGP's strength in managing more complex datasets that involve heterogeneous data and multi-hop relationships. The drug-go-disease network, which includes interactions between drugs, go terms, and diseases, benefits significantly from the diffusion mechanism and meta-path-based techniques employed by GDNDGP.

These results suggest that meta-path-based methods (like HSGCLRDA and GDNDGP) are particularly effective for complex, heterogeneous datasets like drug-go-disease. GNN-based methods perform well on simpler datasets such as DGIdb 4.0 but tend to underperform in more intricate datasets. GDNDGP consistently demonstrates superior performance across all evaluation metrics, with particularly strong results in balanced classification (as shown by MCC) and negative case identification (as indicated by Specificity and NPV). This comprehensive excellence across metrics suggests that GDNDGP's combination of homogeneous and heterogeneous graph aggregation with diffusion-based negative sample generation creates a robust and reliable model for both simple and complex drug-related prediction tasks.

\begin{table}[ht]
    \tiny
    \caption{Performance comparison of methods on the DGIdb 4.0 dataset}
    \label{table:performance1}
    \centering
    \begin{tabular}{l c c c c c c c c}
    \toprule
    \textbf{Models} & \textbf{AUC} & \textbf{AUPR} & \textbf{Recall} & \textbf{Precision} & \textbf{F1} & \textbf{MCC} & \textbf{Spec.} & \textbf{NPV} \\
    \midrule
    \multicolumn{9}{l}{\textbf{GNN-based}} \\
    LightGCN & 0.8128 & 0.8417 & 0.7608 & 0.7885 & 0.7549 & 0.7358 & 0.7367 & 0.7465 \\
    LAGCN & 0.8388 & 0.8547 & 0.7769 & 0.7776 & 0.7768 & 0.7527 & 0.7834 & 0.7655 \\
    MF & 0.8105 & 0.8504 & 0.7518 & 0.7739 & 0.7467 & 0.7496 & 0.7428 & 0.7325 \\
    AGAEMD & 0.8527 & 0.8467 & 0.7996 & 0.8072 & 0.7984 & 0.7635 & 0.7288 & 0.7195 \\
    MNGACDA & 0.8598 & 0.8640 & 0.7737 & 0.7731 & 0.7734 & 0.7634 & 0.7845 & 0.7734 \\
    SGCLDGA & 0.8921 & 0.9663 & 0.8373 & 0.9466 & 0.8786 & 0.8692 & 0.8531 & 0.8372 \\
    \midrule
    \multicolumn{9}{l}{\textbf{meta-path based}} \\
    MGP-DDA & 0.9061 & 0.9674 & 0.9029 & 0.9550 & 0.9078 & 0.8523 & 0.8359 & 0.8226 \\
    HSGCLRDA & 0.9340 & 0.9743 & 0.9132 & 0.9170 & 0.9140 & 0.8635 & 0.8623 & 0.8745 \\
    GDNDGP & \textbf{0.9531} & \textbf{0.9816} & \textbf{0.9357} & \textbf{0.9384} & \textbf{0.9310} & \textbf{0.9045} & \textbf{0.9108} & \textbf{0.8966} \\
    \bottomrule
\end{tabular}
\end{table}

\begin{table}[ht]
\tiny
    \caption{Performance comparison of methods on the drug-go-disease dataset}
    \label{table:performance2}
    \centering
    \begin{tabular}{l c c c c c c c c}
    \toprule
    \textbf{Models} & \textbf{AUC} & \textbf{AUPR} & \textbf{Recall} & \textbf{Precision} & \textbf{F1} & \textbf{MCC} & \textbf{Spec.} & \textbf{NPV} \\
    \midrule
    \multicolumn{9}{l}{\textbf{GNN-based}} \\
    LightGCN & 0.7111 & 0.7487 & 0.8049 & 0.7625 & 0.7581 & 0.7231 & 0.7492 & 0.7385 \\
    LAGCN & 0.7403 & 0.7802 & 0.8661 & 0.8228 & 0.7637 & 0.7412 & 0.7634 & 0.7523 \\
    MF & 0.7812 & 0.8045 & 0.8851 & 0.8327 & 0.7736 & 0.7326 & 0.7512 & 0.7425 \\
    AGAEMD & 0.7711 & 0.7642 & 0.8658 & 0.8222 & 0.7635 & 0.7587 & 0.7723 & 0.7612 \\
    MNGACDA & 0.7905 & 0.7743 & 0.8959 & 0.8427 & 0.7839 & 0.7634 & 0.7845 & 0.7734 \\
    SGCLDGA & 0.8201 & 0.7942 & 0.9062 & 0.8531 & 0.7937 & 0.8543 & 0.8867 & 0.8721 \\
    \midrule
    \textbf{meta-path based} \\
    MGP-DDA & 0.9351 & 0.9437 & 0.8412 & 0.8866 & 0.8601 & 0.8823 & 0.9012 & 0.8912 \\
    HSGCLRDA & 0.9243 & 0.9649 & 0.9037 & 0.9071 & 0.9045 & 0.8932 & 0.9023 & 0.8945 \\
    GDNDGP & \textbf{0.9689} & \textbf{0.9674} & \textbf{0.9166} & \textbf{0.9157} & \textbf{0.9386} & \textbf{0.9127} & \textbf{0.9256} & \textbf{0.9189} \\
    \bottomrule
    \end{tabular}
\end{table}

\begin{figure}[ht]
    \centering
    \includegraphics[clip,scale=0.3]{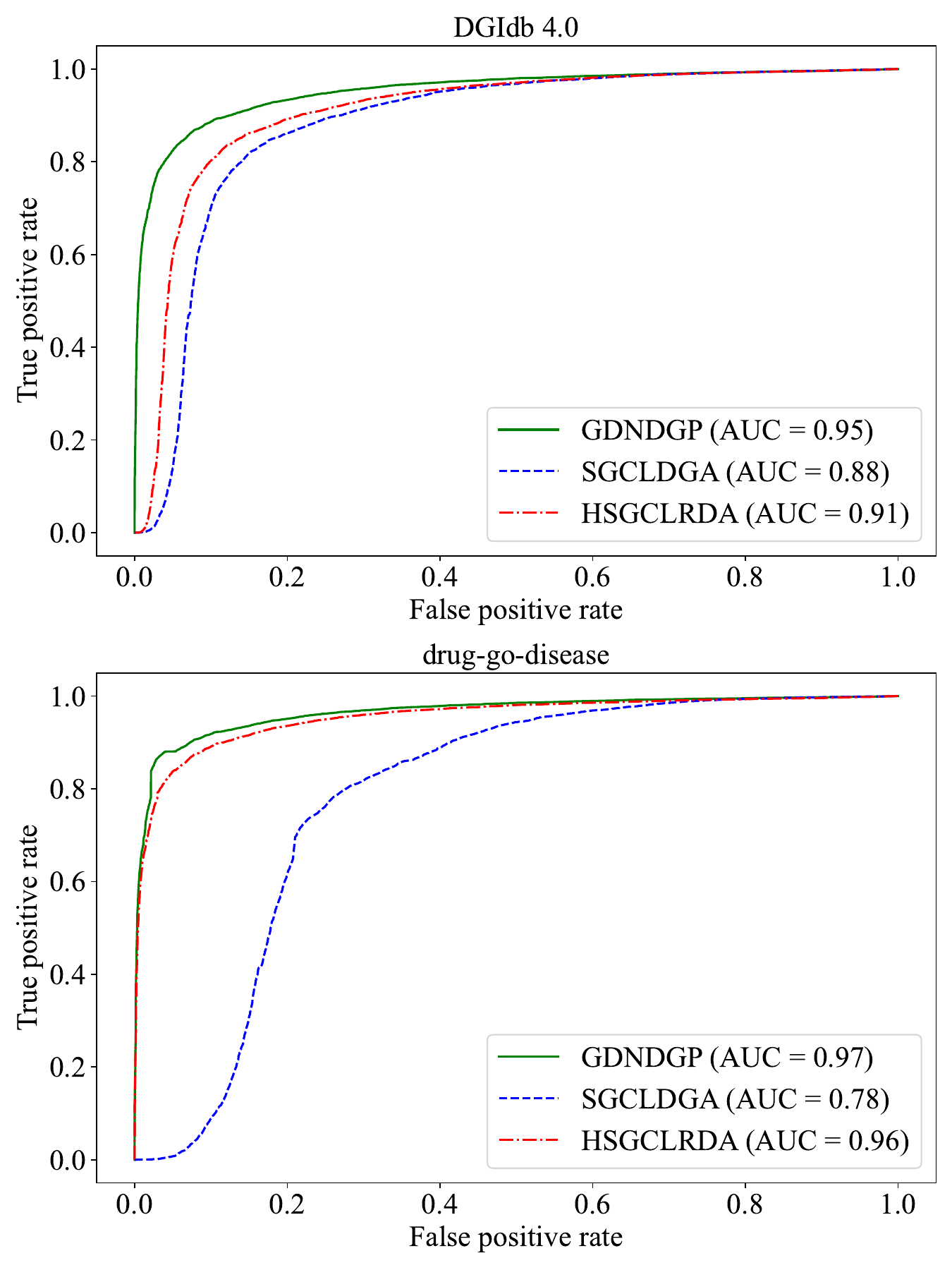}
    \caption{The ROC-AUC comparisons between two datasets.}
    \label{fig:graphs1}
\end{figure}

\subsection{Ablation study (Q2)}

We performed an ablation study on the DGIdb 4.0 dataset to evaluate the contributions of each component in the GDNDGP. The results, presented in Table \ref{table:ablation_study}, demonstrate how the removal of different components affects model performance. The full GDNDGP model achieves the best overall results, with an AUC of 0.9531 and an AUPR of 0.9816, showcasing the strength of combining all modules. When the diffusion mechanism was excluded, the model's AUC dropped to 0.9225, which represents a 3.22\% decrease, indicating that the diffusion mechanism plays a critical role in generating diverse and useful negative samples, helping the model in distinguishing drug-gene pairs more effectively.

Next, we analyzed the impact of removing the homogeneous and heterogeneous graph components. Without the homogeneous graph, the model's AUC declined to 0.9352, resulting in a 1.88\% reduction. Similarly, the removal of the heterogeneous graph component caused a drop in the AUC to 0.9284, which corresponds to a 2.59\% decrease. These results imply that both graph components contribute positively to the overall performance, with each playing a crucial part in capturing the intricate relationships between drugs and genes. However, the diffusion mechanism shows the most significant impact on the model’s effectiveness.

We also compared different strategies for aggregating negative samples generated during the reverse diffusion process in GDNDGP. As shown in Table \ref{table:negative_samples}, we evaluate three aggregation methods: summing, averaging, and applying a weighted sum to the negative samples. These strategies reflect how the generated negative embeddings are combined to improve the model’s prediction performance. The sum strategy aggregates the negative embeddings by adding them together. This approach yielded an AUC of 0.9507, a slight decrease of 0.25\% compared to the full GDNDGP model. While effective, this method did not fully capture the nuances of each sample, as treating all embeddings equally limits its ability to emphasize more informative samples. The average strategy, where the negative embeddings are averaged, led to a more significant performance reduction, with an AUC of 0.9391, reflecting a 1.47\% decrease. This decline indicates that averaging fails to capture the most critical information from the negative samples, as it distributes the importance equally across embeddings, diminishing the contribution of highly informative samples.

The weighted sum strategy, which assigns different weights to each negative sample based on its importance, achieved the highest AUC of 0.9531. By carefully weighting the negative samples, this approach emphasizes embeddings from more important time steps and ensures that the most relevant information is captured. Thus, the weighted sum approach provides the best balance, successfully combining the most relevant information from negative samples and delivering superior results for drug-gene interaction prediction.

\begin{table}[ht]
\centering
\caption{Ablation study results on GDNDGP components}
\label{table:ablation_study}
\begin{tabular}{lcccccc}
\toprule
\textbf{Model}               & \textbf{AUC}  & \textbf{AUPR} & \textbf{Recall} & \textbf{Precision} & \textbf{F1-score} \\ 
\midrule
GDNDGP                        & 0.9531        & 0.9816        & 0.9357          & 0.9684             & 0.9310            \\ 
- diffusion                   & 0.9225        & 0.9585        & 0.8950          & 0.8921             & 0.8935            \\ 

- homogeneous          & 0.9352        & 0.9671        & 0.9123          & 0.9084             & 0.9103            \\ 
- heterogeneous       & 0.9284        & 0.9623        & 0.9031          & 0.8997             & 0.9014            \\ 
\bottomrule
\end{tabular}
\end{table}

\begin{table}[ht]
\centering
\caption{Comparison of different negative sample combination strategies}
\label{table:negative_samples}
\begin{tabular}{lcccccc}
\toprule
\textbf{Combination}              & \textbf{AUC}  & \textbf{AUPR} & \textbf{Recall} & \textbf{Precision} & \textbf{F1-score} \\ 
\midrule
sum                               & 0.9507        & 0.9783        & 0.9240          & 0.9205             & 0.9222            \\ 
average                                    & 0.9391        & 0.9702        & 0.9150          & 0.9115             & 0.9133            \\ 
weighted sum            & 0.9531        & 0.9816        & 0.9357          & 0.9384             & 0.9310           \\ 
\bottomrule
\end{tabular}
\end{table}

\begin{figure}[ht]
    \centering
    \includegraphics[clip,scale=0.3]{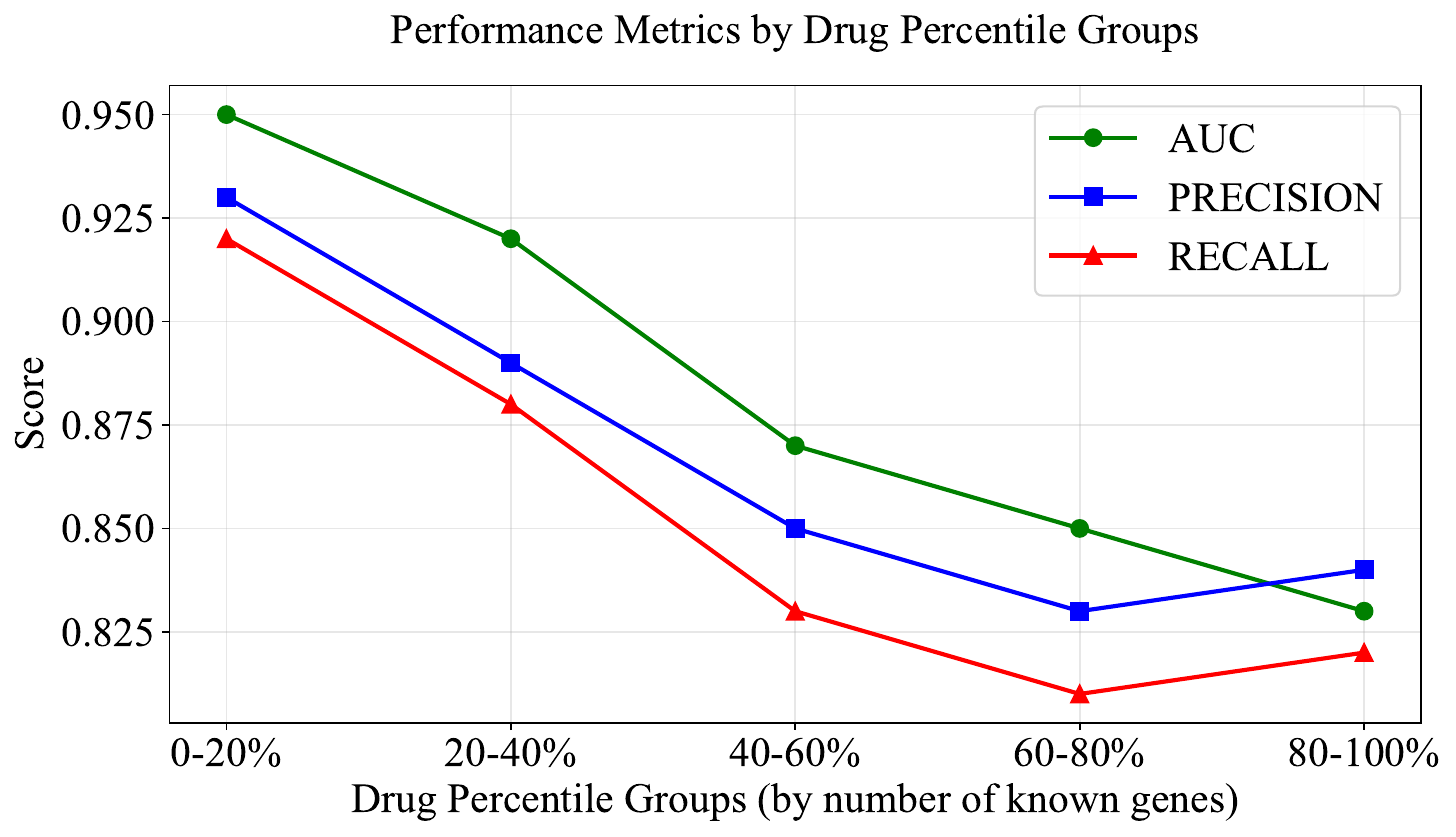}
    \caption{Performance of GDNDGP across drugs with varying numbers of known gene interactions.}
    \label{fig:graphs}
\end{figure}

To better understand how GDNDGP performs across different types of drugs, we analyzed its prediction performance based on the number of known gene interactions for each drug. We divided the drugs into five percentile groups, where the 0-20\% group represents drugs with the highest number of known gene interactions, while the 80-100\% group contains drugs with the fewest known interactions. As shown in Fig. \ref{fig:graphs}, GDNDGP achieves the best performance for drugs with more known interactions, with the 0-20\% group showing the highest scores (AUC: 0.95, precision: 0.93, recall: 0.92). Performance gradually decreases as we move towards drugs with fewer known interactions, with the 80-100\% group showing relatively lower but still reasonable performance (AUC: 0.83, precision: 0.84, recall: 0.82). This pattern suggests that while GDNDGP performs better with well-studied drugs, it maintains acceptable prediction capability even for drugs with limited known interactions, demonstrating its robustness across different data scenarios.

\begin{figure}[ht]
    \centering
    \includegraphics[clip,scale=0.32]{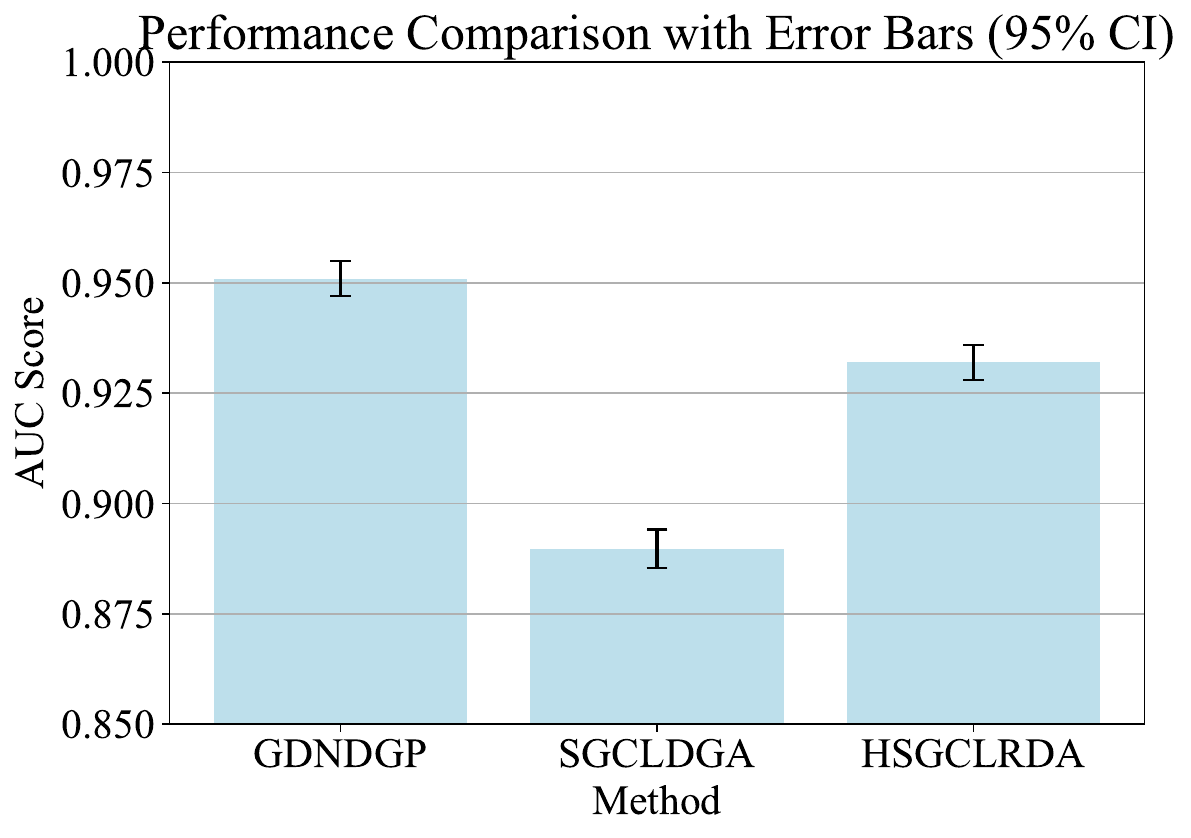}
    \caption{Performance comparison with error bars (95\% CI).}
    \label{fig:statistical}
\end{figure}

Fig. \ref{fig:statistical} illustrates a statistical comparison between GDNDGP and two of the best-performing baseline models: SGCLDGA (a GNN-based method) and HSGCLRDA (a meta-path-based approach). The comparison was based on the mean AUC scores and their corresponding 95\% confidence intervals (CIs) computed across multiple experimental runs to assess both performance and stability. GDNDGP demonstrates superior predictive performance, achieving a mean AUC of 0.953, with its confidence intervals ranging from 0.947 to 0.955. This indicates both high accuracy and consistent performance across different runs. In contrast, SGCLDGA attained a mean AUC of 0.892, with confidence intervals ranging from 0.885 to 0.894, while HSGCLRDA achieved a mean AUC of 0.934, with confidence intervals ranging from 0.928 to 0.936. The significantly narrower confidence intervals for GDNDGP suggest that its results are highly reproducible, underscoring the model’s stability and reliability in drug-gene interaction prediction tasks. These findings reinforce GDNDGP's effectiveness not only in terms of achieving superior absolute performance but also in demonstrating robust and stable results across multiple experimental runs. This comprehensive statistical evaluation highlights GDNDGP as a more reliable and accurate method compared to existing state-of-the-art models for drug-gene interaction prediction.

\subsection{Hyper-parameter analysis (Q3)}

We conducted experiments on the DGIdb 4.0 dataset. We conducted a detailed analysis of two key hyper-parameters in the GDNDGP model: the neighbor threshold $\tau$ and the number of diffusion steps $T$. Figures \ref{fig:graphs1} illustrate how variations in these parameters impact the AUC scores on two datasets. In Fig. \ref{fig:graphs1} (a), we explore the effect of the neighbor threshold $\tau$ on the model’s performance. As shown, the AUC score on DGIdb 4.0 improves as the neighbor threshold increases from 10 to 40, where it reaches a peak value of 0.96. Beyond this point, further increases in the threshold do not yield any significant performance gain. For drug-go-disease, the model performs optimally with a neighbor threshold of 30, achieving its highest AUC score of 0.97. The AUC remains stable with minimal fluctuations as the threshold increases beyond 30.

Fig. \ref{fig:graphs1} (b) investigates the influence of the number of diffusion steps on AUC performance. In DGIdb 4.0, the AUC increases steadily as the diffusion steps rise from 10 to 50, with the highest AUC score of 0.96 obtained at 50 diffusion steps. After this point, increasing the number of diffusion steps provides little to no improvement. Similarly, in drug-go-disease, the AUC reaches a maximum of 0.97 at 50 diffusion steps, and additional steps do not enhance performance further.

These results suggest that on two datasets, an optimal range of neighbor threshold and diffusion steps exist, beyond which performance improvements are negligible. Specifically, a neighbor threshold of 40 and diffusion steps of 50 appear to be the optimal settings for DGIdb 4.0, while a threshold of 40 and diffusion steps of 100 work best for drug-go-disease.

\begin{figure}[ht]
    \centering
    \includegraphics[clip,scale=0.28]{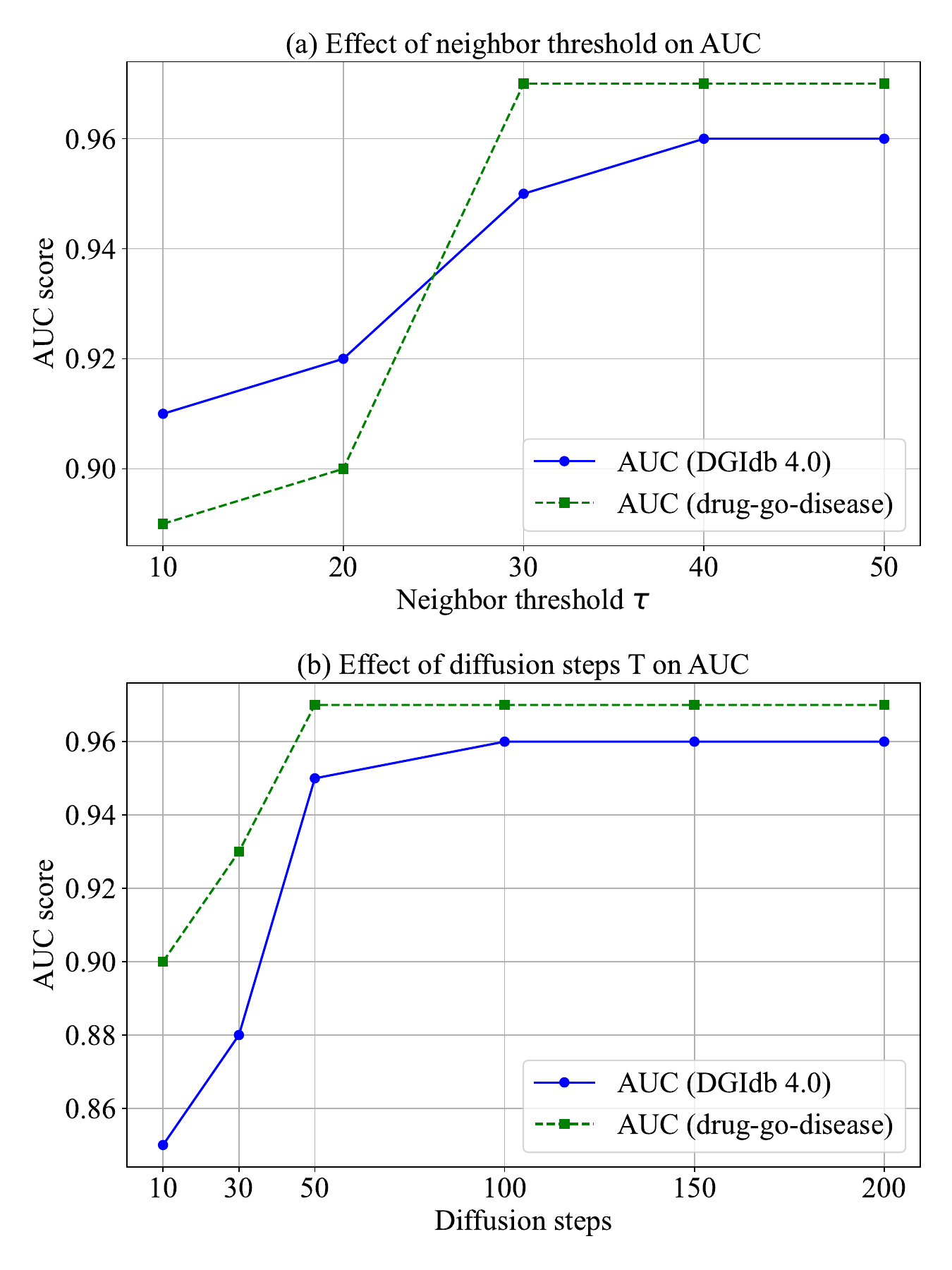}
    \caption{Effect of neighbor threshold $\tau$ and diffusion steps on AUC scores on two datasets.}
    \label{fig:graphs1}
\end{figure}

\subsection{Time analysis (Q4)}
In this section, we conducted experiments on the DGIdb 4.0 dataset. We analyze the effect of two critical hyperparameters embedding dimension and batch size on both model accuracy and computation time, highlighting the trade-offs between performance and efficiency.

\begin{figure}[ht] 
    \centering 
    \includegraphics[clip,scale=0.26]{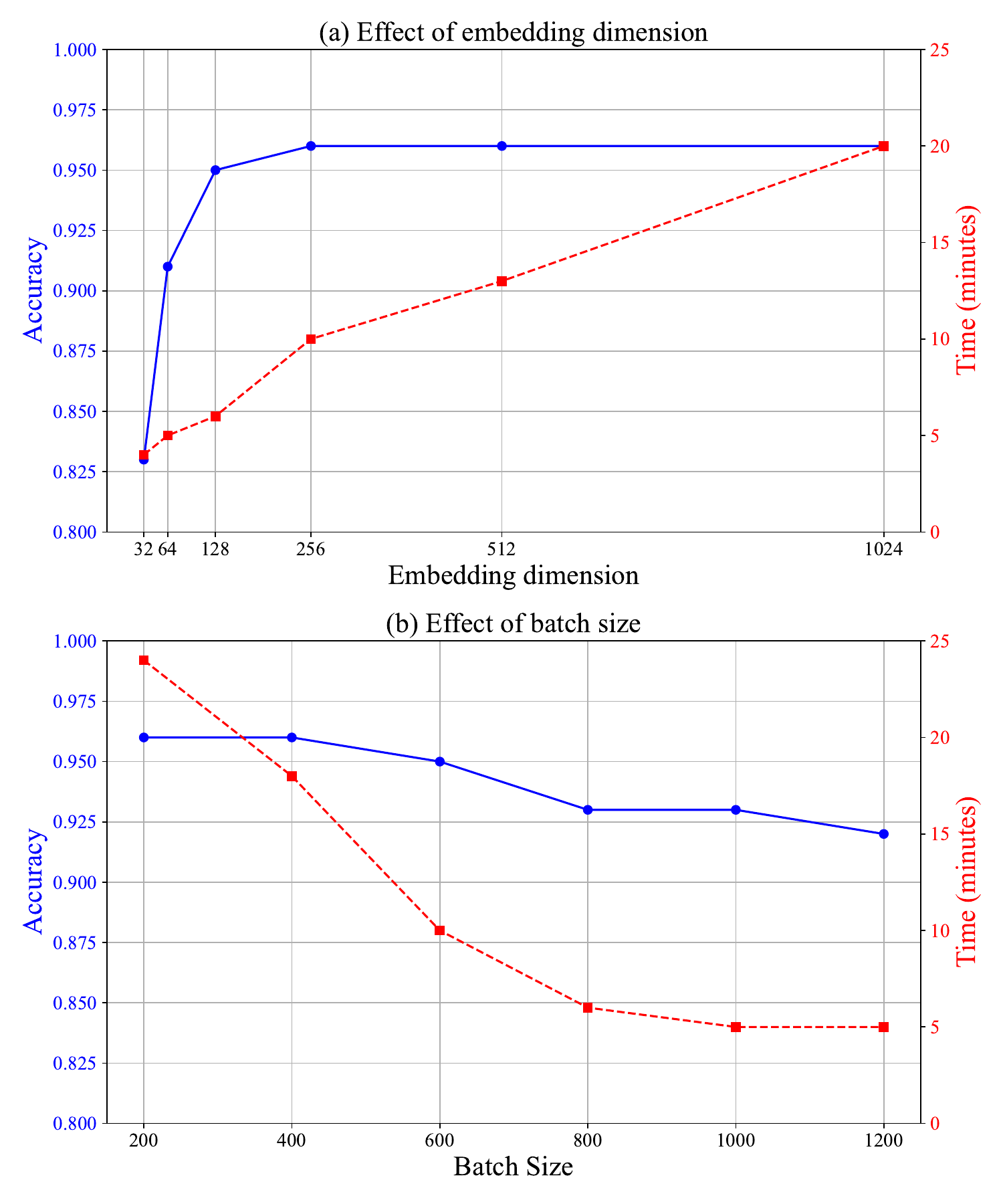} 
    \caption{Effect of embedding dimension and batch size on accuracy and training time.} 
    \label{fig:5} 
\end{figure}

As shown in Fig. \ref{fig:5} (a), we observe that increasing the embedding dimension from 32 to 1024 has a noticeable effect on both accuracy and computation time. The accuracy improves significantly from 0.83 to 0.96 when increasing the embedding dimension from 32 to 128. However, beyond 128 dimensions, further increases in embedding dimension (up to 1024) yield negligible improvement in accuracy, which remains stable at 0.96. On the other hand, the computation time grows substantially as the embedding dimension increases. With an embedding dimension of 32, the model takes approximately 4 minutes to train, while an embedding dimension of 1024 increases the time to 20 minutes. This indicates that, while embedding dimensions beyond 128 do not significantly boost accuracy, they incur a heavy time cost, showing a clear diminishing return in performance versus computation time.

Fig. \ref{fig:5} (b) illustrates the effect of batch size on accuracy and computation time. As the batch size increases from 200 to 1200, accuracy remains stable at 0.96 up to a batch size of 600. However, further increases in batch size result in a slight drop in accuracy, which decreases to 0.92 when the batch size reaches 1200. In contrast, larger batch sizes significantly reduce the computation time. For instance, training time drops sharply from 24 minutes at a batch size of 200 to only 5 minutes at a batch size of 1200. This suggests that while larger batch sizes can improve training efficiency, they may also lead to reduced accuracy if the batch size becomes too large.

There is a clear trade-off between accuracy and computation time for both embedding dimension and batch size. An embedding dimension of 128 strikes a good balance, offering near-maximal accuracy (0.96) with a moderate training time (6 minutes). For batch size, values between 400 and 600 maintain high accuracy (0.96 to 0.95) while keeping the training time reasonable (10 to 18 minutes).

\subsection{Case study}
In this subsection, we focused on evaluating the performance of GDNDGP across various disease categories, specifically anti-cancer drug-gene associations, lung cancer, breast cancer, and less studied subjects, as shown in Table \ref{table:disease}. The objective was to assess the effectiveness of these algorithms in predicting drug-gene associations across diverse contexts.

For the anti-cancer drug-gene association analysis, we used a set of 82 anticancer drugs \cite{fan2024sgcldga}. The predictive performance of SGCLDGA and GDNDGP was compared using the AUC metric. GDNDGP achieved an AUC of 0.9893, which represents a significant improvement of 3.63\% over SGCLDGA's AUC of 0.9548. This indicates GDNDGP's superior capability in predicting drug-gene associations for anti-cancer drugs, which is crucial for advancing targeted cancer therapies.

In the context of lung cancer, the study focused on drugs like Afatinib and genes like BRAF, both of which are highly relevant in lung cancer therapy. The AUC for GDNDGP was 0.9257, representing an improvement of 8.35\% compared to SGCLDGA's AUC of 0.8543. This improvement demonstrates GDNDGP's enhanced ability to predict lung cancer-specific drug-gene associations, making it a more reliable tool for identifying potential therapeutic targets for lung cancer treatment.

For breast cancer, the analysis looked at drugs such as Brivanib and genes like FGFR1, which play critical roles in breast cancer treatment. GDNDGP achieved an AUC of 0.8731, representing a significant improvement of 16.69\% over SGCLDGA's AUC of 0.7482. This result highlights GDNDGP's strong performance in identifying drug-gene associations critical for breast cancer precision medicine.

The experiments also evaluated the algorithms' performance on less frequently studied drugs and genes, including rare drugs like ACRIDINE and genes like A2M. GDNDGP achieved an AUC of 0.9264, representing an improvement of 16.39\% over SGCLDGA's AUC of 0.7961. This finding underscores GDNDGP's robustness and adaptability in predicting associations involving less studied drugs and genes, which is important for uncovering new therapeutic possibilities in under-researched areas.

The experimental results indicate that GDNDGP consistently outperforms SGCLDGA across all tested disease categories, with notable improvements in AUC values. These results validate the effectiveness of GDNDGP and its superior predictive power in various disease contexts, making it a valuable tool for drug-gene association studies. The improvements in AUC across both well-studied and less-studied domains enhance its utility in real-world applications.

\begin{table}[h]
    \centering
    \caption{Comparison of SGCLDGA and GDNDGP on AUC for disease categories}
    \begin{tabular}{lcc}
    \toprule
    \textbf{Disease} & \textbf{SGCLDGA} & \textbf{GDNDGP} \\
    \midrule
    Anti-cancer drug-gene & 0.9548 & 0.9893 \\
    Lung cancer & 0.8543 & 0.9257 \\
    Breast cancer & 0.7482 & 0.8731 \\
    Less studied subjects & 0.7961 & 0.9264 \\
    \bottomrule
    \end{tabular}
    \label{table:disease}
\end{table}

\section{Conclusion} \label{sec: conclusion}
To address the challenges in biomedical research, we introduced GDNDGP, a novel framework that combines homogeneous and heterogeneous graph learning with a diffusion process to generate diverse negative samples for drug-gene interaction prediction. Our approach integrates meta-path construction to enhance message passing between drug-drug and gene-gene, facilitating more effective information exchange. GDNDGP adopts a graph diffusion network to produce hard negative samples, improving the model’s generalization ability by gradually introducing more challenging examples during training. This approach eliminates the need to retrieve all negative samples while achieving high performance. Extensive experiments conducted on two datasets—DGIdb 4.0 and drug-go-disease demonstrate that GDNDGP outperforms state-of-the-art methods in terms of AUC, AUPR, Recall, Precision, and F1-score. Specifically, GDNDGP achieves significant improvements over GNN-based and meta-path-based approaches, especially in heterogeneous datasets. These results highlight the model’s ability to capture both direct and indirect relationships in multi-hop associations, making it highly effective for both well-studied and less-studied drug-gene relationships.

In future work, we will explore several promising directions to enhance the model's capabilities. First, we plan to incorporate molecular fingerprints, drug similarity, and gene function as biological features for initialization to further improve the model's performance \cite{zagidullin2021comparative}. We also want to extend GDNDGP to predict interactions between small molecule drugs and microRNAs (miRNAs) \cite{zhang2010targeting}. The meta-path construction approach in GDNDGP could be adapted to capture the complex relationships between drugs and miRNAs by incorporating features like miRNA expression profiles, secondary structure information, and sequence-based similarities. This extension would leverage GDNDGP's strong performance in handling heterogeneous biological networks while addressing the unique challenges of miRNA-drug interaction prediction, such as the need to consider RNA structural properties and expression patterns.

\bibliographystyle{IEEEtran}
\bibliography{main.bib}

\end{document}